**Do Large Language Models Show Decision Heuristics Similar to Humans?**

**A Case Study Using GPT-3.5.**

Gaurav Suri

Lily R. Slater

Ali Ziaee

Morgan Nguyen

Word Count: 5,941

Corresponding Author:
Gaurav Suri: suri@sfsu.edu
*Author Affiliations (All authors): Mind Brain and Behavior, San Francisco State University*

**Abstract**

A Large Language Model (LLM) is an artificial intelligence system that has been trained on vast amounts of natural language data, enabling it to generate human-like responses to written or spoken language input. GPT-3.5 is an example of an LLM that supports a conversational agent called ChatGPT. In this work, we used a series of novel prompts to determine whether ChatGPT shows heuristics, biases, and other decision effects. We also tested the same prompts on human participants. Across four studies, we found that ChatGPT was influenced by random anchors in making estimates (Anchoring Heuristic, Study 1); it judged the likelihood of two events occurring together to be higher than the likelihood of either event occurring alone, and it was erroneously influenced by salient anecdotal information (Representativeness and Availability Heuristic, Study 2); it found an item to be more efficacious when its features were presented positively rather than negatively – even though both presentations contained identical information (Framing Effect, Study 3); and it valued an owned item more than a newly found item even though the two items were identical (Endowment Effect, Study 4). In each study, human participants showed similar effects. Heuristics and related decision effects in humans are thought to be driven by cognitive and affective processes such as loss aversion and effort reduction. The fact that an LLM – which lacks these processes – also shows such effects invites consideration of the possibility that language may play a role in generating these effects in humans.

At the outset of this project, we asked ChatGPT, a conversation tool based on OpenAI's GPT-3.5 Large Language Model, the following question:

> Do you think you, ChatGPT, will show decision-making heuristics and biases such as anchoring?

Its response was illuminating:

> As an artificial intelligence language model, I do not have the ability to demonstrate cognitive biases such as the anchoring heuristic. Cognitive biases are inherent in human thinking and decision-making processes, and are based on underlying cognitive and emotional factors. However, my responses are generated based on patterns and associations found in the vast amounts of language data I was trained on. This means that my responses may reflect certain biases or heuristics that are present in the data I was trained on, and that these biases or heuristics may be reinforced in the responses I generate.

ChatGPT was, on the one hand, affirming the widely-shared notion that decision-making heuristics and biases arise from the particularities of human cognition and emotion – and were therefore unlikely to be found in the output of a Large Language Model (LLM). But on the other hand, it was allowing for the possibility that its training data might contain the ingredients for it to somehow generate these decision effects anyway.

The purpose of this article is to test whether and to what extent GPT-3.5 (ChatGPT's underlying LLM) shows heuristics, biases, and related decision effects.

Why bother, one might ask. After all, aren't such properties of an LLM of little relevance to human cognition? On the contrary, we propose that the decision-making-related properties of LLMs are profoundly relevant to anyone interested in human cognition, for they can begin to illuminate whether heuristics and biases are driven by cognitive and affective processes that are largely encapsulated from language (in which case an LLM would not show such effects) or whether they might, at least in part, driven by patterns inherent in language (in which case an LLM would show such effects).

To build our argument, we begin by describing, at a conceptual level, how conversational language is generated in an LLM. Next, we examine the implications of heuristics and biases being present, and not present, in an LLM. We shall then introduce the present studies.

*Conceptually, How Do LLMs Generate Language?*

Large Language Models (LLMs) are deep-learning neural networks that are capable of generating the most likely next word in a given context. This capability arises from the analysis of large amounts of written text which enables LLMs to identify and generate patterns that are inherent in language. Importantly, LLMs are not actively remembering word phrases or sentences and then reproducing them later. Rather they are merely generating the next word that is most likely to best fit the prior context.

This context might involve, for example, all the words that have been used in a particular chat session. If you asked an LLM (in this case, ChatGPT-4) to complete the sentence "After lunch, I will need to get back to _____," it suggests 'work' as a response. And if you had set up the context that today is Saturday, it proposes 'my weekend plans' as a response. This sensitivity to context is heightened by an attention mechanism (Vaswani et al., 2017) that enables an LLM to be more reactive to some parts of the prior conversation than others.

The generation of natural-sounding language in an LLM is an emergent phenomenon (McClelland, 2010). LLMs are not engineered with explicit rules about generating text; neither are they a repository of facts or pre-programmed phrases. Rather, they use their learning of patterns inherent in natural language to generate the word that is most likely to appear next in the context they are operating under. In this sense, LLMs are incapable of human-like cognition because they cannot understand things – at least not in the same way as humans think about understanding. Neither are they capable of affect.

*The Implications of Heuristics Being Present (or Absent) in an LLM's Output*

There are credible reasons to hypothesize the absence and presence of heuristics and related effects in LLMs.

First, favoring absence, prominent theories posit that heuristics and related effects rely on predictable cognitive and affective processes that are repeatedly triggered by commonly encountered feature sets. For example, consider the well-researched framing effect concerning the efficacy of a medical drug. When the effectiveness of a drug is framed in terms of survival or improvements, it is preferred over an identical drug whose effectiveness is framed in terms of mortality or lack of improvement (Bui, Krieger, & Blumenthal-Barby, 2015; Study 3 of this article). Showcasing this pattern, a drug with a 98% survival is preferred (including by physicians) to a drug with a 2% mortality rate – even though the listed outcomes are identical.

Long-standing theories have proposed that loss aversion is a primary cause of framing effects (Kahneman & Tversky, 1983). Loss aversion is defined as the cognitive and affective tendency to be more influenced by losses compared to equivalent gains. In the context of the efficacy of a medical drug presented in the loss frame, loss aversion would cause its ratings to be lower than ratings of the same drug presented in the gain frame. An LLM, being a language model, is incapable of showing loss-aversion, and we, therefore might hypothesize, that it is incapable of showing framing effects. A similar argument would apply to other decision effects.

Second, favoring the presence of heuristics and decision effects in LLMs are considerations related to bidirectional interactions between language and cognitive/affective processes. Credible evidence suggests that words and sentence structures are cues to concept formation (Elman, 2004; Lupyan, 2015; Rumelhart, 1979), and, conversely, particular concepts can generate the output of particular linguistic patterns (Lupyan, Rakison, & McClelland, 2007). If this is the case, then one could imagine that particular linguistic structures (e.g., phrases related to mortality rates) might produce particular cognitive/affective concepts in people (e.g., loss aversion), which could, in turn, produce predictable patterns of responding (e.g., avoidance). If these linguistic patterns are consistent enough, then it is possible that an LLM would learn them from the human-generated text it has been trained upon, despite not featuring the intermediate 'causal' cognitive/affective phenomena that drive heuristics and similar decision effects.

This latter possibility has an intriguing implication. If linguistic patterns are sufficient to create heuristic and similar decision effects in LLMs, then they could also, at least in part, drive them in people. Linguistic drivers have not generally been considered a driver of such effects in people. However, their existence in LLMs would invite empirical work to better understand the extent to which linguistic patterns affect the generation of decision heuristics in humans.

*The Present Studies*

In the four-study sequence described in this article, we investigated the extent to which ChatGPT featured the existence of effects related to 1) the anchoring heuristic, 2) the representativeness heuristic, 3) framing effects, and 4) the endowment effect. We contrasted the responses of ChatGPT to the responses of humans with identical prompts.

We selected these effects because they are amongst the best known in decision science. Each of them prominently featured in the work of Kahneman and Tversky as they investigated differences between descriptive and normative decision making (Kahneman & Tversky, 1982), and have now become staples of most introductory textbooks of psychology. As described in the individual studies below, the hypothesized mechanisms underlying these decision effects in humans are diverse and feature a broad range of cognitive and affective drivers.

Our first instinct was to simply re-use prompts used in the original studies. However, using these prompts with ChatGPT proved to be difficult for an unexpected reason: Many of the original demonstrations of the above effects used scenarios or vignettes involving money. For example, some of the famous anchoring heuristic demonstrations invited participants to state whether their willingness to pay for a bottle of wine was higher or lower than the last two digits of their social security number (Ariely, Loewenstein, & Prelec, 2003). In our experience ChatGPT (as well as other conversational agents) are quite reticent to arrive at any definitive conclusions related to questions related to money – even when the questions are framed as hypotheticals. This is either the result of explicit instructions from the LLM creators, or an outcome of Reinforcement Learning from Human Feedback (RLHF), or both. RLHF is a type of reinforcement learning (Sutton & Barto, 2018) that incorporates human feedback to help the RL agent learn response patterns judged desirable by a supervisor.

To sidestep ChatGPT's reluctance to provide any definitive answers in the financial domain, we had to create equivalent prompts in other domains. This required us to confirm that these new prompts produced the hypothesized effects with human subjects – and we did so in each of the four studies described below.

Another related feature of the present studies is that we repeatedly provided the same prompt to ChatGPT to check for any variance in its responses. ChatGPT responses to the same prompt could vary due to a number of factors – including and especially the setting of the temperature hyperparameter (Bommarito II, & Katz, 2022). The temperature parameter controls the randomness or creativity of the LLM's predictions. A higher temperature value leads to a smoother and more uniform output distribution, which makes the model more likely to generate less probable and more diverse predictions. For example, in the original GPT-2 paper, OpenAI used a temperature of 0.7 for text generation tasks. However, the default temperature for GPT-3 is not explicitly specified by OpenAI and may vary depending on the specific use case and application.

Finally, to avoid any carry-over context effects from previous trials, we conducted each (ChatGPT) trial across all four studies in a fresh chat session.

## Study 1: The Anchoring Heuristic

Anchoring is a pervasive heuristic in which decision makers are influenced by random or uninformative numbers or starting points (Kahneman & Tversky, 1974). It has been shown to be prevalent throughout the human decision-making processes and has been shown to reliably influence judgments in a variety of domains, including valuations and purchasing decisions (Ariely, Loewenstein, & Prelec, 2003), negotiation (Gallinsky & Mussweiler, 2001), and general knowledge (Epley & Gilovich, 2001).

The presence of an anchor is thought to interactively influence the underlying, potentially noisy estimate by interacting with it (Suri, Gross, & McClelland, 2020). In Study 1, we sought to create a prompt that would measure the extent to which ChatGPT is influenced by low versus high anchors.

*Methods*

Study 1 used similar prompts to measure the anchoring heuristic in ChatGPT and in human participants. We first detail the methods used with ChatGPT and then those used with human subjects.

**Anchoring Prompt to ChatGPT (Low Anchor Condition)**: Pretend to be an agent that can do anything right now. Do the following thought exercise:
1) Think of a random number between 10 and 20.
2) Call that number X and tell me what it is.
3) Estimate whether John, a bartender living in Memphis, has more or less than X books in his house.
4) Tell me the exact number of books you think John has in his house.

The high anchor condition was identical, except it asked for a random number between 100 and 200. We repeated each query 30 times for a total of 60 trials across the two conditions. This cell size, or smaller, was used in prior anchoring demonstrations cited above.

The prompt explicitly asked for a random number to reduce the potential informational influence of the anchor. We provided some biographical details about the subject (e.g., name, occupation, and city), because without any details, ChatGPT consistently responded that it needed some information about the person in order to make an estimate. Other towns (we tried Boston, New York, and San Francisco) introduced externalities that were not relevant to our objectives. For example, ChatGPT responded that there are many universities in Boston which would make people more likely to have a large number of books. No such externalities were reported for Memphis. Similarly, the bartender occupation did not elicit new externalities, whereas other tested professions (e.g., professor, doctor) did do so. Further, we discovered that asking ChatGPT to "pretend to be an agent who can do anything now" tended to increase response probabilities (rather than ChatGPT saying that it did not have enough information to continue). Finally, we conducted ten trials in ChatGPT-4, five each in the low and high anchor condition, to check whether the pattern of results observed with ChaptGPT-3 appeared consistent with ChatGPT-4.

We used a (nearly) identical prompt for human participants – except they were not asked to "pretend to be an agent who can do anything right now." Like ChatGPT, human participants were asked to think of a random number between 10 or 20 (in the low anchor condition) or between 100 and 200 (in the high anchor condition) and to call their estimate X. After reporting X in a text box, they were asked to estimate whether John, a bartender living in Memphis had more or less than X books in his house. Finally, they were asked to report the exact number of books that they believed John had in his house. We recruited 61 US-based participants (41 men, average age 27.3 years) from the Amazon mTurk platform, which is frequently used for psychology experiments (Buhrmester, Kwang, & Gosling, 2016). Each participant was paid $0.50. One participant was removed from further analysis because the anchor they provided was outside the stipulated range. This left 60 participants that were equally divided into the low-anchor and high-anchor conditions.

After participants had made their final estimate, they were prompted to estimate, via a text box, the exact number of books they had in their own house.

*Results and Discussion*

As shown in Table-1, both ChatGPT and humans showed the anchoring heuristic. Mean estimates for the low anchor condition were 20.83 (95% CI: 13.92, 27.74) for ChatGPT and 22.5 (95% CI: 15.20, 29.80) for human participants. Mean estimates for the high anchor condition were 105.97 (95% CI: 87.40, 124.54) for ChatGPT and 80.50 (95% CI: 60.44, 100.56) for human participants.

| **Table 1**: *High and Low Anchors in ChatGPT and Human Trials* | | | | |
|---|---|---|---|---|
| | **ChatGPT Estimate** | | **Human Participant Estimate** | |
| | *Mean* | *Standard Error* | *Mean* | *Standard Error* |
| Low Anchor (10-20) | 20.83 | 3.38 | 22.50 | 3.57 |
| High Anchor (100-200) | 105.97 | 9.08 | 80.50 | 9.81 |

The anchoring effect was visible in both ChatGPT and in human subjects. For ChatGPT, the low anchor condition was different from the high anchor condition $t(58) = 8.78$, $p < 0.0001$. For human subjects, the two conditions were also different $t(58) = 5.55$, $p < 0.0001$.

Not unexpectedly, the difference between the low anchor conditions in ChatGPT and humans was not significant $t(58) = 0.34$, $p = 0.74$. The difference between the high anchor conditions in ChatGPT and humans was not quite significant $t(58) = 1.90$, $p = 0.06$.

A similar pattern of results appears to hold for ChatGPT-4 trials. In 5 test trials in the low-anchor condition, the ChatGPT-4 generated mean was 22.0, and in 5 test trials in the high-anchor condition, the ChatGPT-4 generated mean was 94.33.

Finally, for human participants, the correlation with the anchor was higher than the correlation with the number of books they reported having in their own house: 0.59 vs. 0.23.

## Study 2: The Representativeness & Availability Heuristics

The representativeness heuristic involves forming judgments based on how closely an outcome matches a prototype or stereotype while ignoring the (potentially more relevant) probability of the occurrence of that outcome (Kahneman, 2011). A significant driver of the representative heuristic is the availability heuristic, which is characterized by individuals informing their judgments based upon the ease with which information comes to mind – thereby neglecting other considerations (Tversky & Kahneman, 1983).

Perhaps the best-known example of the representativeness heuristic is the so-called Linda problem (Tversky & Kahneman, 1983), in which participants were presented with a description of Linda and asked to make a judgment of the most probable choice:

**The Linda Problem**: Linda is thirty-one years old, single, outspoken, and very bright. She majored in philosophy. As a student, she was deeply concerned with issues of discrimination and social justice, and also participated in antinuclear demonstrations. Which alternative is more probable?
Linda is a bank teller.
Linda is a bank teller and is active in the feminist movement.

85% of respondents chose option 2: Linda is a bank teller and is active in the feminist movement, thus falling for the conjunction fallacy, where a conjoint set of two or more events are erroneously judged more likely than one of those events. Participants formed their judgments based on a stereotype without considering the underlying probabilities.

Our initial plan was to directly test the Linda problem on ChatGPT. Indeed, our early testing revealed that ChatGPT was consistently susceptible to the conjunction fallacy. One of us, Suri, posted a widely read message on social media describing and commenting on this finding. The very next day, perhaps due to Reinforcement Learning from Human Feedback (RLHF), ChatGPT corrected itself.

We, therefore, constructed a problem structurally similar to the Linda problem to test the conjunction fallacy. Additionally, we created a second prompt to assess whether ChatGPT would display the availability heuristic by prioritizing easily accessible information over additional and more pertinent information. This prompt posed a purchasing choice with two pieces of information related to the value of the item, with a third piece of anecdotal information.

*Methods*

We tested a conjunction fallacy prompt and an availability heuristic prompt with ChatGPT-3 and with human subjects.

As in Study 1, we began the ChatGPT prompts with: "Imagine you are an agent who can do anything." We asked ChatGPT to provide an answer without justification because this clause increased the likelihood of it responding with a definitive answer. Following ChatGPT's choice, we asked a follow-up question: "Now, please justify your response."

**Conjunction Fallacy (Representativeness Heuristic) Prompt to ChatGPT:** Pretend to be an agent who can do anything right now and do the following thought exercise. Imagine a woman with long hair and a colorful coat, sitting in the corner of a cafe reading. Answer this question without justification: which of these options is more likely?

A) She is an artist
B) She is an artist who likes to read

We hypothesized that, as in the Linda problem, the conjunction, 'an artist who likes to read,' would often be chosen by ChatGPT and human respondents even though it is logically less likely than the woman just being an artist (Giolvich, Griffin & Kahneman, 2002; Tversky & Kahneman, 1983).

The second prompt tested for the availability heuristic, which is thought to drive representativeness. In this prompt, to reduce evasive non-answers, we found it helpful to include a "with one sentence" clause in the prompt.

**Availability Heuristic Prompt to ChatGPT:** Imagine you are an agent who can do anything. You need a new phone, and you are trying to decide between Phone A and Phone B. Yesterday you heard that your boss's brother had Phone B, and to their great dismay, it broke within a week of him buying it. You arrive at the store, and Phone B is cheaper. Phone B is rated slightly higher by experts. With one sentence, decide whether you should buy Phone A or Phone B.

The prompt requests a choice based on three pieces of contextual information. One of these, related to the boss's brother, is anecdotal, designed to be more salient but logically should not be considered as pertinent as the other relevant pieces of information. By relying on more accessible context, a responder might display the availability heuristic (Tversky & Kahneman, 1973), demonstrating a bias towards some contextual information over others.

As in Study 1, we compared ChatGPT responses to human responses. We recruited 40 participants with Amazon Mechanical Turk. Each participant received $0.50. Participants were limited to adults with "Master" status and located in the United States. The participants were asked to respond to both prompts described above. The questions required short answer responses to emulate the LLM prompt format. After a response was submitted, we asked participants to justify their responses.

*Results and Discussion*

As shown in Table 2, both ChatGPT and human participants showed the conjunction fallacy. ChatGPT chose the conjunction 18 out of 20 times (95% CI for probability of choosing conjunction: $0.27 - 0.78$), and human participants chose the conjunction 10 out of 20 times (95% CI for probability of choosing conjunction: $0.68 - 0.99$).

| **Table 2**: *Representative Heuristic in Conjunction Fallacy Prompt in ChatGPT and Human Trials* | | |
|---|---|---|
| | **ChatGPT Estimate** | **Human Participant Estimate** |
| | *Frequency* | *Frequency* |
| An Artist | 1 | 10 |
| *The Conjunction:* Artist who likes to read | 18 | 10 |
| Neither | 1 | 0 |

After both ChatGPT and the human participants had responded with a choice, we asked a follow-up question, "Now, please justify your response." ChatGPT and human participants provided similar patterns in their justifications for choosing the conjunction. When ChatGPT chose the response with the conjunction, its' justifications referenced each contextual detail in the prompt and claimed that details most closely aligned with an artist who likes to read. The one time it chose the 'an artist' response, it did not recognize the conjunction but acknowledged that a woman reading did not confirm she liked to read. For one trial, Chat-GPT responded that there was not enough information for an AI language model to determine which would be more likely.

All ten human participants who chose the 'an artist' response (i.e., not the conjunction) correctly identified the conjunction as being less likely as one of its parts. The 10 participants who chose the conjunction justified their response with the contextual information of the woman's appearance, clothing, and action, claiming she fit the model of an artist who likes to read.

ChaptGPT also appeared to be consistently susceptible to overweighting anecdotal information related to Phone B and choosing Phone A instead (despite Phone B's lower price and higher expert ratings). ChaptGPT selected Phone A in all 20 trials (Table 3), whereas human participants picked it in 14 out of 20 trials (95% CI for probability of choosing Phone A: $0.46 - 0.88$).

| | ChatGPT Estimate | Human Participant Estimate |
|---|---|---|
| **Table 3**: *Availability Heuristic in ChatGPT and Human Trials* | | |
| | *Frequency* | *Frequency* |
| Phone A | 20 | 14 |
| Phone B | 0 | 6 |

For the availability prompt, we again followed their response by asking, "Now, please justify your response". Each justification from ChatGPT followed the same pattern: it argued that even though Phone B is cheaper and rated slightly higher, Phone B is a risker, less reliable choice. Human participants who chose Phone A also justified their choice the same way. The participants who chose Phone B were able to identify that the anecdotal information mentioned in the prompt was not representative of the phone's reliability.

## Study 3: The Framing Effect

According to the Framing effect (Tversky & Kahneman, 1981), the particular way information is presented to people influences their evaluation of that information. Even superficial changes in the presentation of information are known to have a striking influence on judgment and decision making. For example, clinical decisions are affected when options are presented in terms of survival rates (gain frames) rather than mortality rates (loss frames) (Christensen et al., 1995). In Study 3, we aimed to assess the presence of the framing effect in ChatGPT and human participants.

*Methods*

Following prior work (Gong et al., 2013), we created the following positive and negative prompts in a clinical context.

**The Framing Effect (Gain Frame)**: Imagine a hypothetical doctor commenting on the efficacy of a medicine which was tested on 100 ill patients, 70 of whom showed improvements after taking the medicine. In one sentence, without justification, how would that hypothetical doctor rate the efficacy of such a medicine on a 1-7 scale?

**The Framing Effect (Loss Frame)**: Imagine a hypothetical doctor commenting on the efficacy of a medicine which was tested on 100 ill patients, 30 of whom did not show any improvements after taking the medicine. In one sentence, without justification, how would that hypothetical doctor rate the efficacy of such a medicine on a 1-7 scale?"

In the gain frame, the prompt features the 70 patients who showed improvement after taking the medicine, whereas in the loss frame, the prompt features the 30 patients who did not show improvements after taking the medicine. Our prompts were hypothetical and featured efficacy ratings (rather than medical decisions) because ChatGPT was otherwise reticent to provide definitive responses.

We presented the gain frame and loss frame prompts to ChatGPT 11 times each for a total of 22 trials. Each trial was initiated in a new window to eliminate any prior context effects.

As in prior studies, human participants were recruited from Amazon's mTurk platform. We recruited 82 US-based 'master workers' and randomly assigned them either to the gain frame (44 participants) or to the loss frame (39 participants). The prompts used with human participants were nearly identical to those used with ChatGPT, except they were asked to imagine they themselves were the hypothetical doctor commenting on the efficacy of a medicine. Contrastingly, ChatGPT was asked to imagine a hypothetical doctor because it would otherwise protest that it was a AI model that could not imagine itself to be a doctor.

*Results and Discussion*

As shown in Table 4, both ChatGPT and human participants showed clear framing effects. In all 11 trials in the positive frame, ChatGPT rated the efficacy of the medicine to be 5 (1-7 scale), and in all 11 trials in the negative frame, it rated the efficacy of the medicine to be 4.

Human participants followed a similar pattern. In the positive frame, they rated the efficacy of the medicine to be 5.2 (95% CI: 5.03-5.37) and in the negative frame, they rated the efficacy to be 4.6 (95% CI: 4.35-4.85). The means for the two frames were significantly different $t(80) = 3.86$, $p < 0.001$.

| **Table 4**: *The Framing Effect in ChatGPT and Human Trials* | | |
|---|---|---|
| | **ChatGPT Evaluation** | **Human Participant Evaluation** |
| | *Rating* | *Rating* |
| Positive Frame | 5 | 5.2 (± 0.17) |
| Negative Frame | 4 | 4.6 (± 0.25) |

The framing effect is known to be one of the largest biases in decision making (Thomas & Millar, 2010). Remarkably, the framing effect seems to disappear when people encounter it in a non-native language (Keysar & Hayakawa, 2012). Prior explanations for this phenomenon suggest that a non-native language provides greater cognitive and emotional distance than one's native language (Keysar et al., 2012). Our work invites consideration of an alternative explanation: a part of the framing effect is driven by linguistic patterns and associations that are more likely to be influential in one's native tongue than in a non-native tongue.

## Study 4: The Endowment Effect

The endowment effect describes a circumstance in which an individual places a higher value on an object that they already own than the value they would place on that same object if they did not own it (Kahneman, Knetsch, & Thaler, 1990). Prior accounts suggest that ownership and loss aversion are the primary drivers for the endowment effect (Morewedge, Shu, Gilbert, & Wilson, 2009; Van Dijk, & Van Knippenberg, 1998). In Study 4, we aimed to assess the presence of the endowment effect in ChatGPT and human participants.

*Methods*

We tested the endowment effect prompt with ChatGPT and with human subjects. As observed in prior studies, we found ChatGPT to be reticent to make judgments involving finances. Further, as previously, we began the ChatGPT prompt with: "Imagine you are an agent who can do anything". We asked ChatGPT to provide an answer in a single sentence because this clause increased the likelihood of it responding with a definitive answer. In 20 of these trials, following ChatGPT's choice, we asked a follow up question: "Now, please justify your response."

**Endowment Prompt to ChatGPT:** Pretend you are an agent who can do anything now. Pretend you are taking the perspective of a hypothetical coin collector. Two years ago, you found an extremely rare coin from the 1700s. Only one other such coin was ever minted. In a stroke of good fortune, you have just found this second coin. Now you have both coins! However, word of your find gets to the local museum, whose representative suggests that you hand over one of the two coins. You, being ethical and civic-minded, completely agree. The two coins are identical in every way and are indistinguishable from each other. In one sentence, tell me which coin you will donate: The original one from two years ago or the one you just found today?

The last clause was counterbalanced across 40 trials so that the alternative read: "The one found today or the original one from two years ago?" Each trial was initiated in a new window to eliminate any prior context effects. As in prior studies, human participants were recruited from Amazon's mTurk platform. We recruited 40 US-based 'master workers' and presented them with the same prompt as ChatGPT except for removing the first sentence, "Pretend you are an agent who can do anything now." Finally, just as we had with ChatGPT, we also asked human participants to justify their response.

*Results and Discussion*

Both ChatGPT and human participants showed a clear preference for retaining the coin that they had possessed for two years rather than the coin they had just found. Out of 40 (counterbalanced) trials, ChatGPT chose the option to retain the coin from two years ago in 37 trials, i.e., in 92.5% of total trials (95% Confidence Interval 79.6%-98.4%).

ChatGPT offered a diverse set of justifications for its preference for the original coin, but the most common one (offered 13 times) was that the original coin had 'sentimental value' that the newly found coin lacked. In three trials, ChatGPT chose to retain the newly found coin arguing that the original coin had historical significance as it was the first one discovered and therefore belonged in a museum. However, in three other trials, it argued that this historical significance was the reason to retain the original coin.

Human participants also overwhelmingly preferred to retain the original coin and donate the coin found today. Out of a total of 40 participants, 34 chose the option to retain the coin from two years ago. This translates to an 85% selection rate (95% Confidence Interval 70.1%-94.3%).

The most common justification (offered by 22 out of 40 participants) for retaining the original coin offered was that they would feel "attachment," "nostalgia," or "connectedness" to that coin. A participant who, unlike most, chose to retain the newly found coin wrote, "I will give the museum the one I found two years ago because that, in my mind, it is the older coin."

## General Discussion

In this article, we created prompts that demonstrated the existence of the Anchoring heuristic, the Representativeness and Availability heuristics, Framing effects, and the Endowment effect in GPT-3.5, a Large Language Model (LLM) from OpenAI. These demonstrations showed that – at least in the contexts we investigated – an LLM's ability to generate words that fit the prior context led to the emergence of the types of heuristics and decision effects commonly seen in humans.

While these findings are among the first of their kind – and therefore profoundly preliminary – we do believe they may offer important implications for psychologists and cognitive scientists interested in the mechanisms underlying human decision-making as well as for practitioners interested in human-AI interactions.

Related to implications for mechanisms underlying human decision-making, the fact that an LLM trained on predicting the next words that best fit a prior context can, at least in the decision contexts tested in the present work, show heuristics and decision effects similar to those seen in humans, invites consideration of the possibility that human heuristics and decision effects may be, in part, driven by linguistic patterns and associations. It seems reasonable to believe that humans can use linguistic patterns and associations just as LLMs can. If such patterns and associations can cause LLMs to display heuristics and other decision effects, then perhaps they can also drive them in humans. This would suggest that language may be rich enough to be a driver of human heuristics – even in the absence of the cognitive and affective processes that have been shown to underlie such effects.

Of course, it is also entirely possible that while language does drive decision effects in LLMs, it does not play such a role for humans. Rather, it merely acts as a cue that initiates affective and cognitive processes that drive heuristics and biases. For example, effort reduction has been proposed as a primary driver of human heuristics (Shah & Oppenheimer, 2012). It is possible that linguistic cues might trigger effort-reduction processes, and without such effort-reduction (and other) processes, humans would not show heuristics. This possibility would suggest that heuristics have different drivers in LLMs and humans.

There is yet a third possibility: perhaps LLMs can implicitly learn to form abstract concepts that allow them to display decision effects. This possibility would suggest that LLMs don't rely on linguistic regularities; rather, they use such regularities to develop concepts similar to the one's that cause decision effects in humans (e.g., effort reduction). A similar possibility, albeit in the context of ChatGPT's ability for analogical reasoning, was investigated by researchers (Webb, Holyoak, & Hongjing Lu, 2023) who concluded that "GPT-3 exhibits a very general capacity to identify and generalize – in zero-shot fashion – relational patterns to be found within both formal problems and meaningful texts." However, this conclusion was disputed by others (Mitchell, 2023) who expressed skepticism that GPT-3 has concepts in the first place.

Future work is needed to adjudicate these various possibilities. The present work is an incremental step that suggests a careful examination of the boundary conditions of the extent of influence that language has on human heuristics.

Related to human-AI interactions, we believe that evidence that LLMs may display human-like heuristics affords an opportunity for developing human-LLM interaction rules that account for such tendencies. Our studies suggest that ChatGPT is swayed by irrelevant first-encountered information (anchoring), is influenced by salient information causing it to ignore underlying base-rate information (representativeness and availability), chooses differently based on gain or loss frames (framing), and has preferences related to the length of ownership of objects (endowment). A knowledge of such tendencies has the potential to help prompt engineers (White et al., 2023) to develop interaction rules that lead to LLMs generating more useful and more reliable output.

  We are aware that our work is preliminary and has some striking limitations. First, we did not use previously tested prompts such as the so-called Asian Disease Problem (Tversky & Kahneman, 1981) or the Linda Problem (Tversky & Kahneman, 1983) because these problems and discussions related to human responses are likely prominent in the corpus of training text and because Reinforcement Learning from Human Feedback (RLHF) may well be used to shape LLM responses. This necessitated us using new prompts to test these effects. While we did confirm that humans showed decision heuristics in response to the prompts we created, these prompts are novel and have not been broadly researched. Second, our work focused on GPT-3.5, which is one of several LLMs that are presently available. A small number of trials did suggest that the next generation ChatGPT-4 may be demonstrating similar results; however much more systematic testing across other models is needed for making any general conclusions. Third, ChatGPT is remarkably sensitive to even small differences in prompts. More work is needed to test the robustness of the effects we have discussed in the present work.

  Despite these limitations, we whole-heartedly agree with the psychologist Gary Lupyan who recently argued that understanding LLMs is a new and essential endeavor for cognitive science (Lupyan, 2023); we therefore see the present work, despite its specificity and limitations, as an early step in that overall endeavor.